\documentclass[nohyperref]{article}

\usepackage{amsmath,amsfonts,bm}

\newcommand{\loss}{\mathcal{L}}

\newcommand{\measurements}{y}
\newcommand{\measurementop}{\mathcal{H}}
\newcommand{\evolutionop}{\mathcal{M}}
\newcommand{\prior}{\mathcal{G}}
\newcommand{\state}{u}
\newcommand{\latent}{z}
\newcommand{\parameters}{\theta}
\newcommand{\gnn}{\textrm{\sc gnn}}

\newcommand{\domain}{\Omega}
\newcommand{\uinit}{u_{\text{init}}}

\newcommand{\Secref}[1]{Section~\ref{sec:#1}}
\newcommand{\Figref}[1]{Figure~\ref{fig:#1}}
\newcommand{\figref}[1]{Fig.~\ref{fig:#1}}
\newcommand{\tableref}[1]{Tab.~(\ref{table:#1})}
\newcommand{\Tableref}[1]{Table~(\ref{table:#1})}

\renewcommand{\eqref}[1]{Eq.~(\ref{eq:#1})}

\newcommand{\x}{\textbf{x}}
\newcommand{\uxy}{\textbf{u}}
\newcommand{\uxyinit}{\textbf{u}_{\text{init}}}

\DeclareMathOperator*{\argmin}{arg\,min}
\DeclareMathOperator*{\minimize}{minimize}

\usepackage{microtype}
\usepackage{graphicx}
\usepackage{subfigure}
\usepackage{booktabs} 
\usepackage{hyperref}
\usepackage{url}
\usepackage{multirow}
\usepackage{graphicx}
\usepackage{pifont}
\usepackage{booktabs}
\usepackage{siunitx}
\usepackage{wrapfig}
\graphicspath{ {images/} }
\usepackage{hyperref}

\usepackage[accepted]{icml2022}

\usepackage{amsmath}
\usepackage{amssymb}
\usepackage{mathtools}
\usepackage{amsthm}
\usepackage{multibib}
\newcites{supp}{Supplementary References}
\hypersetup{
    colorlinks=true,   
    urlcolor=blue,
    pdftitle={Overleaf Example},
    pdfpagemode=FullScreen,
}
\newcommand{\beginsupplement}{%
        \setcounter{table}{0}
        \renewcommand{\thetable}{S\arabic{table}}%
        \setcounter{figure}{0}
        \renewcommand{\thefigure}{S\arabic{figure}}%
        \setcounter{section}{0}
        \renewcommand{\thesection}{S\arabic{section}}
        \setcounter{equation}{0}
        \renewcommand{\theequation}{S\arabic{equation}}
     }

\usepackage[capitalize,noabbrev]{cleveref}

\theoremstyle{plain}

\theoremstyle{definition}

\theoremstyle{remark}

\usepackage[textsize=tiny]{todonotes}

\begin{document}

\twocolumn[
\icmltitle{Learning to Solve PDE-constrained Inverse Problems with Graph Networks}



\icmlsetsymbol{equal}{*}

\begin{icmlauthorlist}
\icmlauthor{Qingqing Zhao}{stanford}
\icmlauthor{David B. Lindell}{stanford}
\icmlauthor{Gordon Wetzstein}{stanford}
\end{icmlauthorlist}

\icmlaffiliation{stanford}{Stanford University}

\icmlcorrespondingauthor{Qingqing Zhao}{cyanzhao@stanford.edu}
\icmlcorrespondingauthor{David B. Lindell}{lindell@stanford.edu}
\icmlcorrespondingauthor{Gordon Wetzstein}{gordon.wetzstein@stanford.edu}

\icmlkeywords{Machine Learning, ICML}

\vskip 0.3in
]



\printAffiliationsAndNotice{} 

\begin{abstract}
Learned graph neural networks (GNNs) have recently been established as fast and accurate alternatives for principled solvers in simulating the dynamics of physical systems. 
In many application domains across science and engineering, however, we are not only interested in a forward simulation but also in solving inverse problems with constraints defined by a partial differential equation (PDE).  
Here we explore GNNs to solve such PDE-constrained inverse problems. 
We demonstrate that GNNs combined with autodecoder-style priors are well-suited for these tasks, achieving more accurate estimates of initial conditions or physical parameters than other learned approaches when applied to the wave equation or Navier--Stokes equations. We also demonstrate computational speedups of up to $90\times$ using GNNs compared to principled solvers. Project page: \href{https://cyanzhao42.github.io/LearnInverseProblem}{https://cyanzhao42.github.io/LearnInverseProblem}
\end{abstract}

\section{Introduction}
\label{sec:introduction}

Understanding and modeling the dynamics of physical systems is crucial across science and engineering. Among the most popular approaches to solving partial differential equations (PDEs) are mesh-based finite element simulations, which are widely used in electromagnetism~\citep{pardo2007self}, aerodynamics~\citep{economon2016su2,ramamurti2001simulation}, weather prediction~\cite{bauer2015quiet}, and geophysics~\citep{schwarzbach2011three}. Learning-based methods for mesh-based simulations have recently made much progress~\citep{pfaff2020learning}, offering faster runtimes than principled solvers, better adaptivity to the simulation domain compared to grid-based convolutional neural network (CNNs)~\citep{thuerey:2020,wandel2020learning}, and generalization across simulations. State-of-the-art learning-based mesh simulators operate on adaptive meshes using graph networks~\citep{pfaff2020learning}. While this approach has been successful in simulating the dynamics of physical systems, it remains unclear how to leverage these network architectures to efficiently solve PDE-constrained inverse problems of the form in \eqref{invproblem}.
\begin{equation}
\begin{split}
	\minimize_{ \left\{ \state_0, \parameters \right\} }& \,\, \loss \left( \measurementop \left( \state_{t=1 \ldots T} \right), \measurements_{t=1 \ldots T} \right), \quad \\
	s.t. &\,\, \state_{t+1} = \evolutionop_{\parameters} \left( \state_{t} \right),
	\label{eq:invproblem}
\end{split}
\end{equation}
\begin{figure}
\begin{center}
    \vspace{-0.4cm}
    \includegraphics[width=0.485\textwidth]{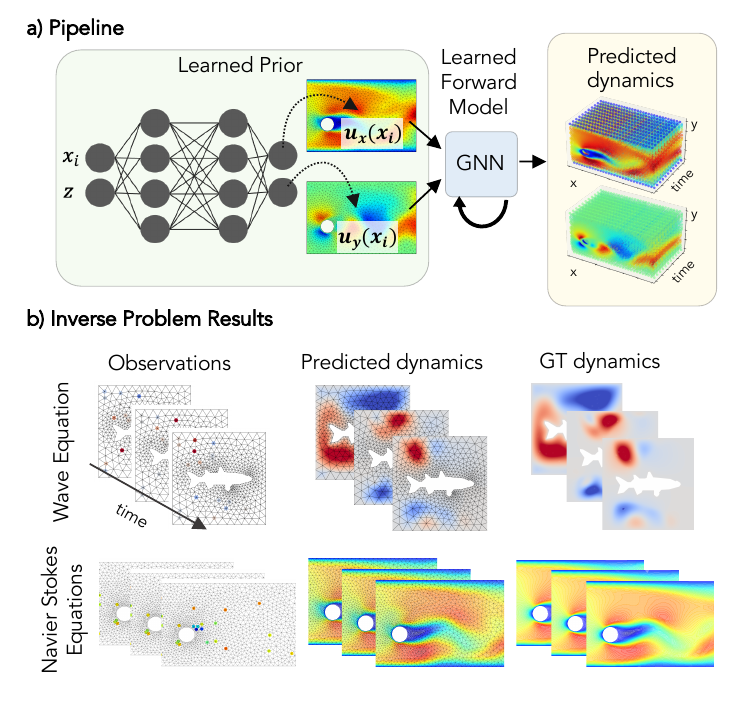}
    \vspace{-0.6cm}
    \caption{\textbf{(a)} Pipeline for our approach: we first pre-train the prior and a Graph Neural Network (GNN) for a given class of problems in a supervised manner using a dataset generated from classical FEM solvers. Here, the GNN is a fully-differentiable forward simulator and the prior is a generative model learned over the space of physics parameters of interest. At test time, the generative model maps the latent code to the estimated physics parameter that is passed to the GNN to obtained the predicted dynamics. The latent code is then optimized to minimize the difference between predicted and observed dynamics on the graph. \textbf{(b)} We demonstrate the proposed method for solving inverse problems governed by the wave equation or Navier--Stokes equations. From sparse observations, the method recovers the complete evolution of the dynamics.}
    \vspace{-0.3cm}
\end{center}
\label{fig:demo}
\end{figure}

where $\state_t \in \mathbb{C}^N$ is the state of the simulation modeled at $N$ nodes of the mesh at time $t=1 \ldots T$; the operator $\measurementop : \mathbb{C}^N \rightarrow \mathbb{C}^M$ models the measurement process of the real or complex-valued state using a set of $M$ sensors, giving rise to the observations $\measurements_t \in \mathbb{C}^M$; $\evolutionop_{\parameters}: \mathbb{C}^N \rightarrow \mathbb{C}^N$ is an operator representing the PDE constraints that govern the temporal dynamics of the system with $\parameters \in \mathbb{R}^P$ parameterizing $\evolutionop$ at time $t$. In the context of inverse problems, we seek the initial state $\state_0$ or the parameters $\parameters$ given a set of observations $y_{t=1 \ldots T}$. These problem are often ill-posed and considerably more difficult than learning the forward simulation.

Here we explore efficient approaches that leverage recently proposed graph networks to solve PDE-constrained inverse problems. For this purpose, we consider learning-based approaches to the simulation, where $\evolutionop$ is modeled by a graph network $\gnn: \mathbb{C}^N \rightarrow \mathbb{C}^N$. 
Moreover, we formulate priors for both initial state $\state_0 = \prior_\state \left( \latent_\state \right)$ and the parameters $\parameters = \prior_\parameters \left( \latent_\parameters \right)$ that constrain these quantities to be fully defined in a lower-dimensional subspace by the latent codes $\latent_\state  \in \mathbb{R}^{L_\state}$, $L_\state<N$, and $\latent_\parameters  \in \mathbb{R}^{L_\parameters}$, $L_\parameters<P$.
The fact that the graph network $\gnn$, and finite element methods in general, operate on irregular meshes motivates us to explore emerging coordinate network architectures~\citep{park2019deepsdf,tancik2020fourier,sitzmann2020implicit} as suitable priors. Coordinate networks operate on the continuous simulation domain and map coordinates to a quantity of interest, such as the initial condition or the parameters of a specific problem. \\\\
To our knowledge, this is the first approach to combining continuous coordinate networks with graph networks for learning to solve PDE-constrained inverse problems. The summary of the proposed approach is shown in \Figref{demo}. We demonstrate that this architecture affords faster runtimes and better quality for fewer observations than the principled solvers we tested, while offering the same benefits of improved accuracy over grid-based CNNs to inverse problems that graph networks offer for forward simulations (\Figref{demo}).

\section{Related Work}
\label{sec:related}
\begin{figure*}[th!]
    \begin{center}
        \includegraphics[width=\textwidth]{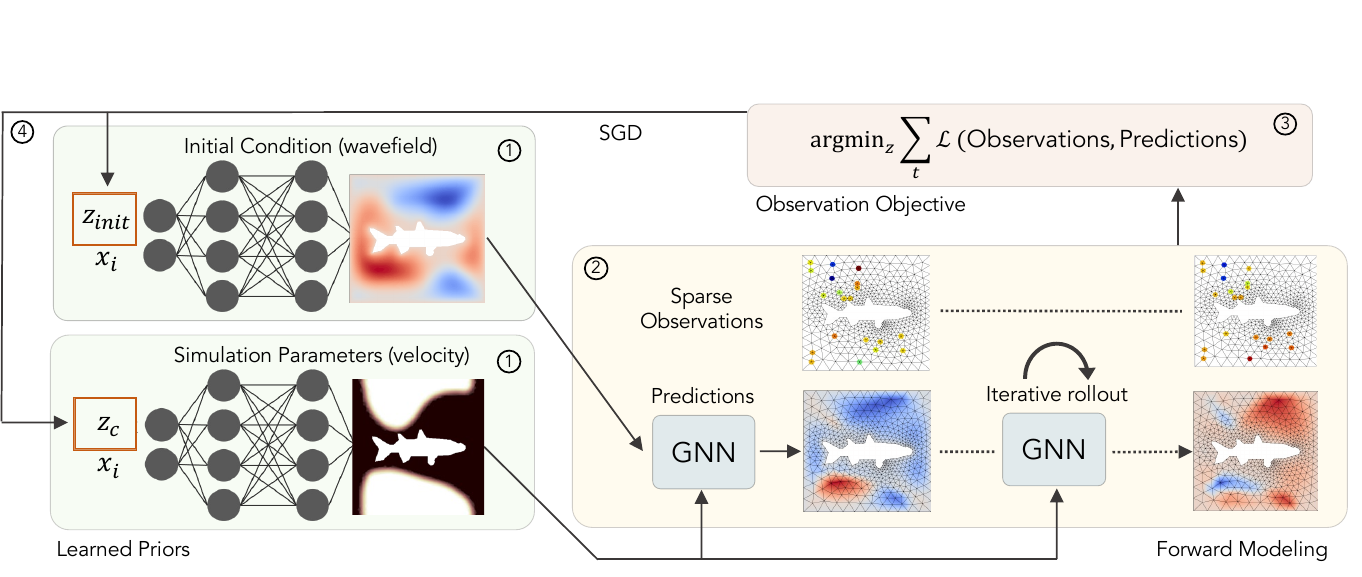}
    \end{center}
    \vspace{-0.5cm}
    \caption{Pipeline for solving inverse problems (illustrated for the 2D wave equation). The forward simulator GNN (the blue boxes) and the prior networks (green boxes) are pre-trained with a dataset of wave trajectories generated using classical FEM solvers. We aim to recover the initial wavefield or the velocity distribution. At test time, the generative model first maps a latent code to the estimated physics parameters (step 1) that is passed to the GNN to obtain the predicted dynamics (step 2). The latent code is optimized to minimize the difference between the predicted and the observed dynamics on the graph (steps 3 and 4).}
    \label{fig:wave_pipeline}
\end{figure*}

Machine learning methods have emerged as a powerful tool for modeling the dynamics of physical systems. Compared to conventional methods (see e.g., ~\citet{bookFEM}), physics-based machine learning techniques may offer improved computational performance, ease of implementation, and learning priors from data to solve ill-posed physics-based inverse problems. Previous techniques span a spectrum of being entirely data-driven (i.e., modeling dynamics with a feedforward pass through a network)~\citep{bhatnagar2019prediction, li2020fourier, tompson2017accelerating,lu2019deeponet,lino2021simulating,pfaff2020learning,sanchez2020learning,seo2019physics} to hybrid approaches that either directly incorporate conventional solvers~\citep{um2020solver,belbute2020combining,bar2019learning,kochkov2021machine} or adapt similar optimization schemes or constraints~\citep{karniadakis2021physics,wang2021long,sirignano2018dgm,raissi2019physics,wang2021learning,wandel2020learning}.

Many of the aforementioned methods operate on regular grids, for example using convolutional neural networks (CNNs). A major disadvantage of this approach is its inability to allocate resolution adaptively throughout the simulation domain. Learning-based finite element methods (FEMs)~\citep{xue2020amortized} and \textit{graph neural networks} (GNNs)~\citep{kipf2016semi,hamilton2017inductive,velivckovic2017graph,wu2020comprehensive,scarselli2008graph} have been shown to be a powerful method for simulation on irregular grids with adaptive resolution. GNNs process data structured into a graph of nodes and edges, and have been demonstrated for modeling complex physical dynamics with particle- or mesh-based simulation~\citep{lino2021simulating,seo2019physics,li2020neural,li2020multipole}. Our approach is inspired by recent GNN architectures for modeling time-resolved dynamics~\citep{pfaff2020learning,sanchez2020learning} that generalize to unseen initial conditions and solution domains. Whereas these approaches focus on learning the forward simulation given a set of initial or boundary conditions, ours aims at solving ill-posed PDE-constrained inverse problems that require a learned simulation model as part of the framework.

Machine learning techniques have recently also been adapted for solving ill-posed physics-based inverse problems. These problems often involve recovering solutions to PDEs~\citep{raissi2020hidden} or physical quantities, such as the density, viscosity, or other material parameters~\citep{mosser2020stochastic,he2021reparameterized,fan2020solving}, from a sparse set of measurements. Other methods can super-resolve PDE solutions from estimated solutions at coarse resolution~\citep{esmaeilzadeh2020meshfreeflownet} or recover initial conditions from PDE solutions observed at coarse resolution later in time~\citep{li2020neural,frerix2021variational}. Yet another class of techniques are tailored to the inverse design problem, which aims to optimize material properties such that the PDE solution satisfies certain useful properties, e.g., for the design of nanophotonic elements~\citep{molesky2018inverse}. Our approach is most similar to other methods that infer material properties or initial conditions from sparse measurements of the PDE solution over time. Similar to some of these methods~\citep{mosser2020stochastic,jiang2019global}, we use a generative model to learn a prior over the solution space of material parameters or initial conditions; however, our framework is the first to learn to map a latent code to material parameters or initial conditions compatible with a GNN forward model. This allows our model to solve physics-based inverse problems on irregular grids with adaptive resolution, leading to improved computational efficiency and accuracy.

\section{Method}
\label{sec:method}
We approach the task of solving physics-constrained inverse problems by first pre-training a GNN as a fast forward model that outputs solutions to partial differential equations given the initial states and boundary conditions. The physics-constrained inverse problems we aim to solve involve recovering initial conditions or other parameters that govern the evolution of the PDE. Since recovering these quantities is an ill-posed problem, we also learn a prior over the space of solutions using a deep generative model. Here, a network maps latent codes to a space of material parameters or initial conditions on the graph that comprise plausible solutions to the inverse problems. At test time, the GNN is a fully-differentiable forward simulator, and the latent code of the generative model is optimized to minimize the difference between the predicted and the observed dynamics on the graph. A detailed pipeline of our approach for solving inverse problem for the wave equation (\Secref{experiments}) is shown in \Figref{wave_pipeline}.

\subsection{GNN-based Simulator}

GNNs have previously been found to perform well in simulating PDE models \cite{pfaff2020learning,sanchez2020learning}. The excellent performance can be attributed to their adaptive resolution and ability to generalize by modeling complicated local interactions. Here, we leverage the state-of-art GNN \cite{pfaff2020learning} to learn a fast forward model that outputs solutions to partial differential equations given the initial states and boundary conditions.

The state of the system at time $t$ is described as $\state_t=(V,E)$ with nodes $V$ and mesh edges $E$ that define the mesh. We adopt the approach of \cite{pfaff2020learning} to model the temporal dynamics of the system, i.e., $\evolutionop_{\parameters}$, using a GNN with Encoder-Processor-Decoder architecture followed by an integrator.

We encode the relative displacement vector (i.e., the vector that points from one node to another) and its norm as edge features, $e_i$. The node features comprise dynamics quantities that describe the state of the PDE and a one-hot vector, $v_i$, indicating the type of the nodes (domain boundaries, inlet, outlet, etc.). In the forward pass, the encoding step uses an edge encoder multilayer perceptron (MLP) and node encoder MLP to encode edge features and node features into latent vectors. The processing step consists of several message passing layers with residual connections. This step takes as input the set of node embeddings $v_i$, and edge embeddings $e_{ij}$, and outputs an updated embedding $v_i$ and $e_{ij}$. The equations that define the layers of the message passing steps are given by the following.
\begin{align}
    e_{ij}' &= \textrm{\sc mlp}_e(e_{ij},v_i,v_j), \quad  \\
    v_i' &= \textrm{\sc mlp}_v(v_i,\sum_j e_{ij}').
\end{align}

In the final decoding step, an MLP is used to transform the latent node features $v_i$ to the output $p_i$, which we integrate at each time step to update the dynamic quantities $\state_{t+1}=\state_t+p_t$.


\subsection{Learning Inverse Problems and Priors}

The inverse problems we wish to solve are of the form outlined in Equation~\ref{eq:invproblem} with the dynamics being learned by the GNN described above. Solving for the initial condition or parameters may be ill-posed, so we restrict the solutions to lie in a lower-dimensional subspace of the autodecoder-type priors $\prior_\state$ or $\prior_\parameters$. When working on regular grids, these priors could easily be implemented as CNN-based generative model, but it is not clear how to model such priors for irregular meshes. To this end, we leverage recently proposed coordinate networks with ReLU activation functions, which use a conditioning-by-concatenation approach to generalize across signals:
%
\begin{equation}
\begin{split}
		&\prior_{\state/\parameters} = \mathbf{W}_n \left( \phi_{n-1} \circ \phi_{n-2} \circ \ldots \circ \phi_0 \right) \left( \left[ 
		\begin{array}{c}
				x\\
					z_{\state/\parameters}
		\end{array}
		\right] \right) + \mathbf{b}_n \\
		&\phi_i \left( x_i \right) = \textrm{ReLU} \left( \mathbf{W}_i \mathbf{x}_i + \mathbf{b}_i \right),
\end{split}
\label{eq:prior}
\vspace{-0.1cm}
\end{equation}
where $\phi_i: \mathbb{R}^{M_i} \mapsto \mathbb{R}^{N_i}$ is the $i^{th}$ layer of the MLP consisting of the affine transform defined by the weight matrix $\mathbf{W}_i \in \mathbb{R}^{N_i \times M_i}$ and the biases $\mathbf{b}_i\in  \mathbb{R}^{N_i}$. The input to the first layer concatenates the coordinate $x$ with a scene-specific latent code vector $z_{\state/\parameters}$. This coordinate network, implemented by an MLP, maps continuous coordinates on the simulation domain to a quantity of interest. Thus, it can be evaluated on a regular grid or, more importantly, on the irregular locations of the graph nodes our GNN operates on. 

We learn  $\prior_{\state/\parameters}$ in a pre-processing step, using training data relevant for a specific physical problem. We optimize the network parameter, denoted by $\psi_\prior$, and the latent code, $\latent$, with respect to the individual training sample,  indexed by $i$, to maximize the joint log posterior of all training samples:
\begin{align}
    \argmin_{\psi_\prior,z}\sum_{i=1}^N \left(\sum_{j=1}^{K}\loss \left( \prior \left(z^i,\x_j\right),u^i_j \right) +\frac{1}{\sigma^2} \|z^i\|_2 ^2\right)
\label{eq:loss_prior}
\vspace{-0.1cm}
\end{align}

At test time, the pre-trained GNN modeling the dynamics and the pre-trained coordinate networks as the prior are fixed, and the latent code vectors $z_{\state/\parameters}$ are optimized for a given set of sparse observations using a standard Adam solver~\citep{kingma2014adam}. \\

\section{Experiments}
\label{sec:experiments}
We demonstrate our approach on the wave equation and the Navier--Stokes equations for modeling acoustics or fluid dynamics. Our method leverages a learned simulator using a Graph Neural Network and a learned prior for fast and accurate recovery of the unknown parameters. The U-Net (CNN) solver baseline is comparable to existing approaches in the literature for forward simulation \cite{thuerey:2020,wandel2020learning, holl2020learning, bhatnagar2019prediction, tompson2017accelerating}; since we focus on solving physics-constrained inverse problems, we extend this approach accordingly and incorporate our learned prior. We also evaluate the effect of the learned prior. We find that our approach gives a favorable tradeoff between accuracy and speed, enabling accurate recovery of the unknown parameters while being up to nearly two orders of magnitude faster than the classical FEM solver.\\
\begin{figure}
  \begin{center}
    \includegraphics[width=0.32\textwidth]{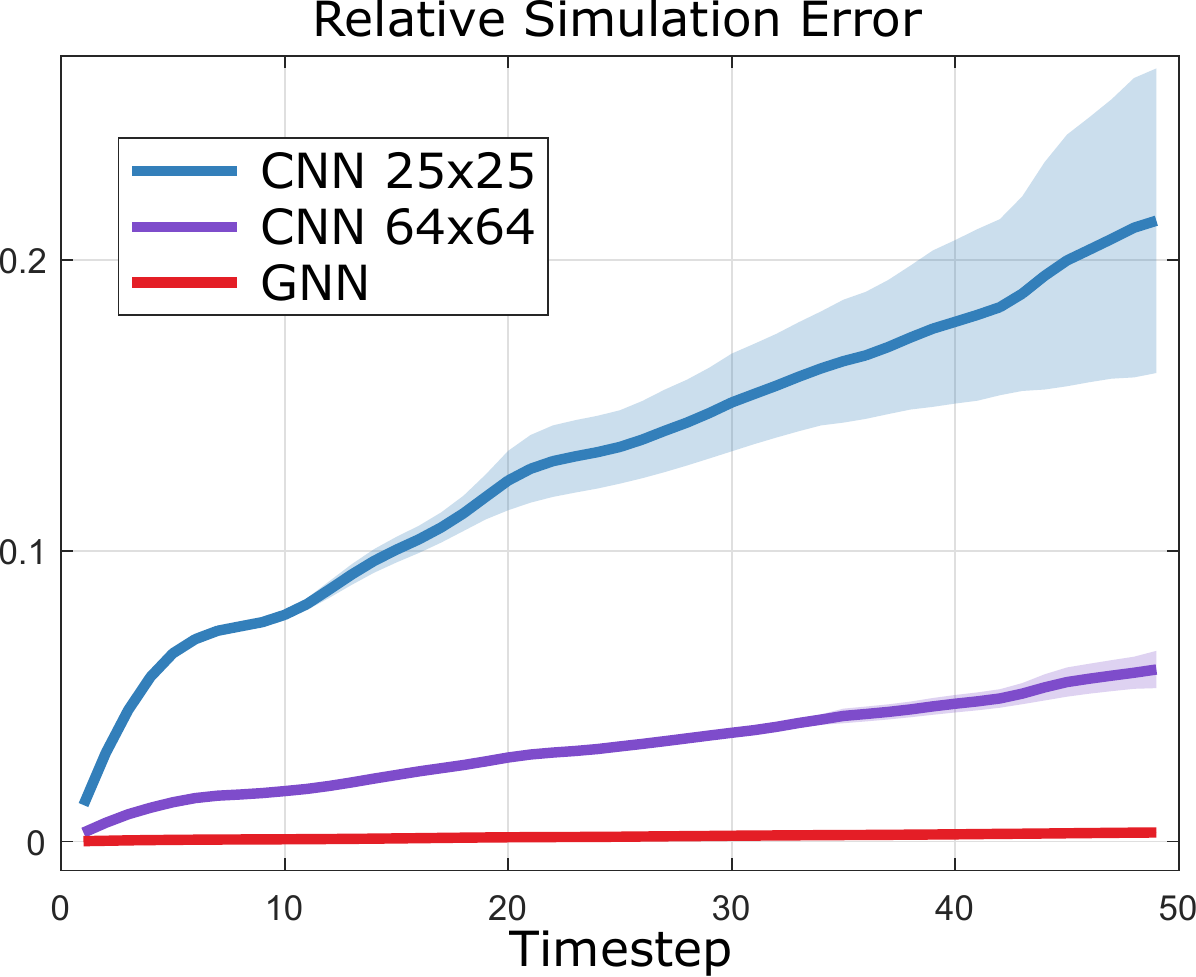}
  \end{center}
  \vspace{-0.1cm}
  \caption{Accumulated relative error of the U-Net (CNN) (on both coarse and fine grids) and GNN-based forward simulators for varying number of time steps.}
  \label{fig:solver_comparison}
  \vspace{-0.1cm}
\end{figure}
\begin{figure*}
  \begin{center}
      \includegraphics[width=\textwidth]{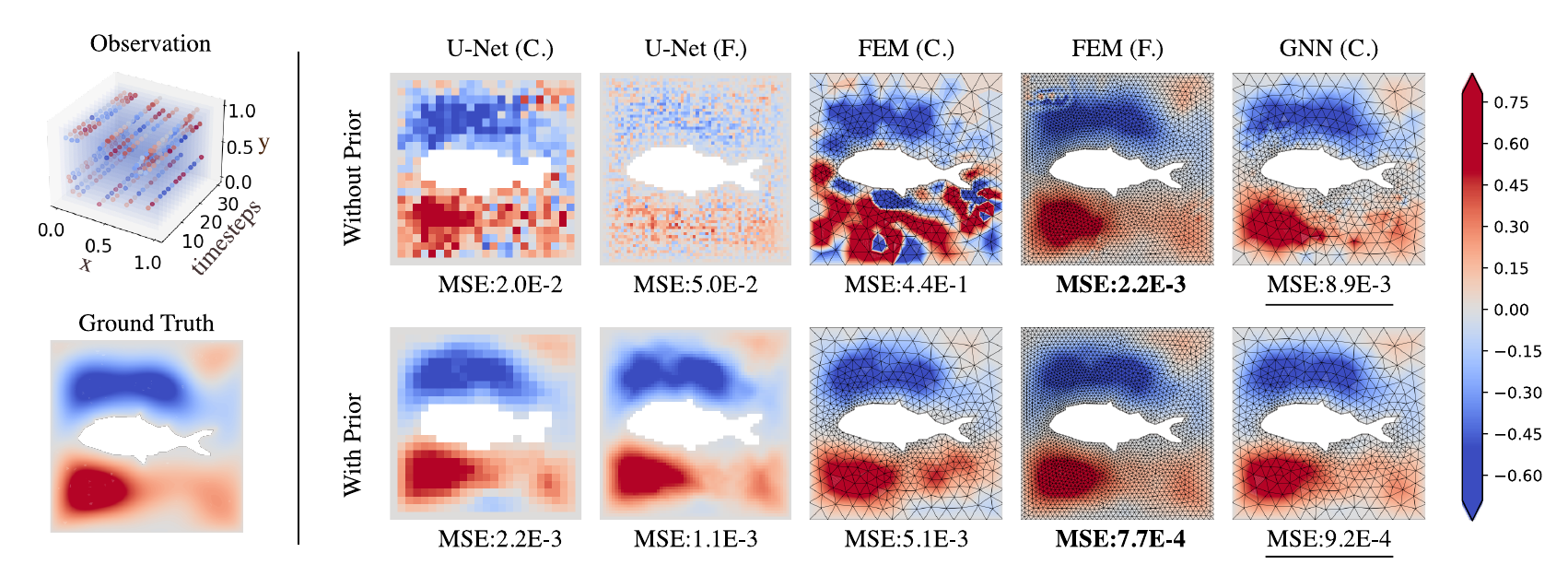}
  \end{center}
  \vspace{-0.3cm}
  \caption{Qualitative results for the initial state recovery problem with the 2D wave equation. Here ``C.'' refers to the coarse meshes ($25\times25$ nodes for U-Net (CNN) or $\approx600$ nodes for GNN), and ``F." refers to the fine meshes ($64\times64$ nodes for U-Net (CNN) or $\approx2800$ nodes for the FEM solver). Without the learned prior, $\prior$, all methods fail due to the ill-posed nature of the problem. Using the prior, we find the GNN yields a result that is closer to the ground truth compared to both U-Net (CNN) models. While the FEM solver on the fine grid outperforms the GNN, it is also $\approx$8$\times$ slower. Thus the GNN with the prior gives a favorable tradeoff between speed and accuracy.}
  \label{fig:wave_init_fig}
\end{figure*}
\subsection{Two-Dimensional Scalar Wave Equation}
The wave equation, \eqref{wave_equation}, is a second-order partial differential equation that describes the dynamics of acoustic or electromagnetic wave propagation. 
\begin{equation}
\begin{split}
		\frac{\partial^2 u}{\partial t^2}-c^2\frac{\partial^2 u}{\partial x^2} &= 0 \quad \\
		u_0' = \frac{\partial u}{\partial t}|_{t=0} &=0 \\		
		u_0 &= \uinit \quad 
\end{split}
\label{eq:wave_equation}
\end{equation}
Here, $u$ is the field amplitude and $c$ is the velocity parameter that spatially characterizes the speed of the wave propagation inside the medium. The wave equation can be solved given initial conditions, $u_0'$ and $\uinit$, and the boundary conditions.  In our experiments, the domain is chosen to be $\domain \in [0,1]^2$ with an obstacle at the center. The Dirichlet boundary conditions are used at all boundaries, $u|_{\partial \domain} = 0$. The initial field velocity is set to be 0, and the initial field distribution is denoted by $\uinit$.

The PDE-constrained inverse problem we are aiming to solve is to recover unknown parameters $\uinit$ or $c$ from sparse measurements of $u$ in space and time. Formally, it can be described as
\begin{equation}
\begin{split}
    \minimize_{ \left\{ \latent \right\} }& \,\,\sum_{t\in T_{\text{iter}}}\|\measurementop(u_t^{\text{gt}}(\x)) - \measurementop(u_t^{\text{pred}}(\x)) \|^2_2 \quad \\
    s.t.&\,\, \underbrace{u_{t+1} = u_{t}+\evolutionop_{c} \left( u_{t},u_{t}' \right)}_{\quad u_0 \left (\x \right )=\prior_{\uinit}(z,\x) \quad \text{or}\quad c\left (\x \right ) = \prior_c(z,\x)}
\label{eq:invproblem_wave}
\end{split}
\end{equation}

Here, we minimize the mean squared error (MSE) between the predicted field ($u_t^{\text{pred}}$) and the ground truth observed field ($u_t^{\text{gt}}$), summed over different observation time steps, $t$. We address two different tasks, (1) recovering the initial condition, $\uinit$, and (2) full-waveform inversion~\citep{virieux2009overview} in which we recover the velocity parameter, $c$. These optimized quantities are parameterized by the prior network $\prior_{\uinit/c}$. Finally, the measurement sampling operator is given by $\measurementop$, and $\evolutionop_c$ is the forward model that solves the wave equation, \eqref{wave_equation}, to produce the field at the next time step.

\paragraph{Learned Forward Simulation}
In our approach, $\evolutionop_c$ is given by a GNN, which learns the wave equation forward model on irregular meshes. 

As a baseline, we train a U-Net (CNN) forward model based on \cite{thuerey:2020}.
Similar to \cite{pfaff2020learning}, we train the GNN and U-Net (CNN) by directly supervising on a dataset of ``ground truth'' wave equation solutions generated with an open source FEM solver (FEniCS~\citep{LoggMardalEtAl2012}).
The FEM solver uses a fine irregular mesh with many more nodes ($\approx$2800) than the GNN ($\approx$600) on average across the dataset. The training dataset is composed of 1100 simulated time-series trajectories using 37 separate meshes. We evaluate on 40 held-out trajectories across 3 different meshes. We find that the FEM solver requires a timestep that is 5$\times$ smaller than the GNN solver to achieve stable results. At this setting, the FEM solver simulated on a fine irregular mesh is roughly 8$\times$ slower than the GNN; running the FEM solver on the same grid as the GNN is 2.5$\times$ slower than the GNN with accuracy roughly 80$\times$ worse in terms of MSE.

To compare the accuracy of different learned forward simulators, we plot the accumulated relative error, unrolled for 50 time steps and averaged over 96 trajectories in \figref{solver_comparison}. The GNN-based simulator provides a robust solution. We evaluate two different cases for the U-Net (CNN). In one case, we restrict the number of grid nodes to be roughly the same as the GNN (25 $\times$ 25). 
Since the small number of nodes must be regularly and coarsely spaced, this provides poor performance. Second, we set the grid resolution of the U-Net (CNN) such that the spacing between grid points matches the minimum spacing in the GNN mesh. This requires 7$\times$ more nodes than the GNN ($64 \times 64$), but the performance is still relatively worse than the GNN. 

\begin{table*}[ht!] 
  \centering
  \begin{tabular}{lc|cc|cc|c}\toprule
  \textbf{Forward Model}   &\textbf{\# nodes}   &\multicolumn{2}{c|}{\textbf{Initial State Recovery}}  &\multicolumn{2}{c|}{\textbf{FWI}}  & Runtimes \\
  & & MSE       & MSE (with $\prior$)     & MSE   & MSE (with $\prior$) &  (s) \\\midrule
  FEM  (Irr. F.)	   		& 2827    & \textbf{1.55e-3}      &\textbf{7.82e-4}       &3.91e-1					    &  1.19e-1               &1.25e1\\
  FEM  (Irr. C.)  	    & 611     &1.72e-1                &4.18e-3                &\underline{3.88e-1} 	& 	1.64e-1		           &6.03e0 \\
  U-Net  (Reg. C.)      & 625     & 1.32e-2               &3.52e-3                &	4.19e-1				  	  & 	1.54e-1 	           &\textbf{1.71e-1} \\
  U-Net  (Reg. F.)		  & 4096    & 3.16e-2               &1.49e-3                & 5.14e-1 				    & \underline{1.08-1}		 &\underline{1.91e-1}\\
  GNN  (Irr. C.)        & 611 	  &\underline{1.05e-2}    &\underline{8.87e-4}    & \textbf{3.68e-1}		& \textbf{1.05e-1}       &1.44e0\\\bottomrule
  \end{tabular} 
  \caption{Quantitative results for the initial state recovery and full-waveform inversion (FWI) problems for the 2D wave equation.  All metrics are averaged over 40 test samples and evaluated on coarse irregular grids. Observation setup: we use measurements every 2 GNN time steps from 2 to 30 GNN time steps with 20 sensors randomly sampled from the grid. Runtime measures the time taken per optimization iteration. We observe that the GNN outperforms the U-Net (CNN) with a similar number of nodes and provides comparable results with the U-Net (CNN) using $7\times$ more nodes. The classical FEM solver using a fine irregular mesh gives the lowest MSE, however it is at least $8\times$ slower than the learned simulator approaches.}
  \label{table:init_state_table}
\end{table*}

\paragraph{Initial State Recovery}
For the initial state recover task, we solve \eqref{invproblem_wave} where $\uinit(\x)=\prior_{\uinit}(\latent,\x)$ is the unknown parameter and initial velocity is given to be 0.
For both this problem and full-waveform inversion, the prior is pre-trained on a dataset of 10,000 values of $\uinit$ (or $c$), generated by sampling a Gaussian random field and tapering the solution to zero near the boundaries to satisfy the Dirichlet boundary condition.

We solve the inverse problems by optimizing the value of $\latent$ using the ADAM optimizer \cite{kingma2014adam} for all experiments until convergence, or a maximum of 2000 iterations.
In the case where we optimize without the prior, we directly optimize the unknown parameters, $\uinit$ or $c$.

\Tableref{init_state_table} reports a quantitative comparison of different solvers with and without the prior, and a qualitative comparison is shown in \figref{wave_init_fig}. Here we unroll the GNN for 30 GNN time steps and take measurements every 2 GNN steps from 2 to 30 time steps. At every measurement, we have sensors placed at 20 randomly sampled nodes from the coarse irregular mesh. For baseline comparisons with other form of meshes (regular for U-Net (CNN) and fine irregular mesh for FEM solver), we use nearest neighbor sampling to find the corresponding sensor location. From the table we observe that with or without the prior, our approach provides a favorable trade-off between the accuracy and the speed. While the U-Net (CNN) is the fastest forward simulator, the GNN solver outperforms the U-Net (CNN) solver in terms of accuracy at resolution $25\times 25$ and performs slightly better than the U-Net (CNN) at resolution $64\times 64$. We attribute this to the stronger performance of the GNN forward model, as shown in \figref{solver_comparison}. Given that the FEM solver operating on the finest mesh gives the most accurate forward model, it gives the best MSE in the inverse problem. However, it is at least $8\times$ slower than the learned simulator approaches. For all cases, using a learned prior significantly improves the final accuracy. By optimizing the latent code, we constrain the solution space to the manifold learned by $\prior_{\uinit/c}$ and avoid bad local optima that lie far outside the dataset distribution.

\paragraph{Full Waveform Inversion}
Full-waveform inversion (FWI) is a common problem in seismology \cite{fwi_review} and involves recovering the density of structures in the propagation medium.
We are motivated by recent work that uses variants of a CNN as a prior~\cite{mosser2020stochastic,wu2019parametric} or forward operator~\cite{fwi_fourier_op} for seismic inversion.
As we are primarily interested in a comparison between learned methods on regular and irregular meshes, we adopt  the U-Net (CNN) baseline as a representative method that can be unrolled per timestep and integrated into our framework.
Since the prior is agnostic to the particular grid setup, we use the same prior across all models.

For the FWI problem, we minimize \eqref{invproblem_wave} where $c(\x)=\prior_c(\latent,\x)$ is the unknown parameter. FWI is a highly non-linear inverse problem, and the final optimization result depends heavily on the initial state and may easily fall into local minima. In order to improve the optimization, different forms of progressive training schemes have been used. For example, one can progressively fit the data in time by introducing a damping function as in \citep{fwi-timedamping}. In frequency-domain full-waveform inversion, one can first optimize low-frequencies to avoid local minima \citep{fwi-frequency,fwi-frequency2}. In our approach, we adopt a progressive training scheme in time where we gradually increase the total number of observed time steps as the optimization proceeds. For the FWI task, we unroll the GNN for 30 GNN time steps and take measurements every 2 GNN steps. At the beginning of the optimization, we only have observations at time step $T=\left\{2\Delta t\right\}$, and we include one extra time step's measurement every 120 optimization iterations until $T=\left\{2\Delta t,4\Delta t,\cdots, 30\Delta t\right\}$.

\begin{figure}[h!] 
  \begin{center}
    \vspace{0.2cm}
    \scalebox{1}[-1]{\includegraphics[width=0.4\textwidth]{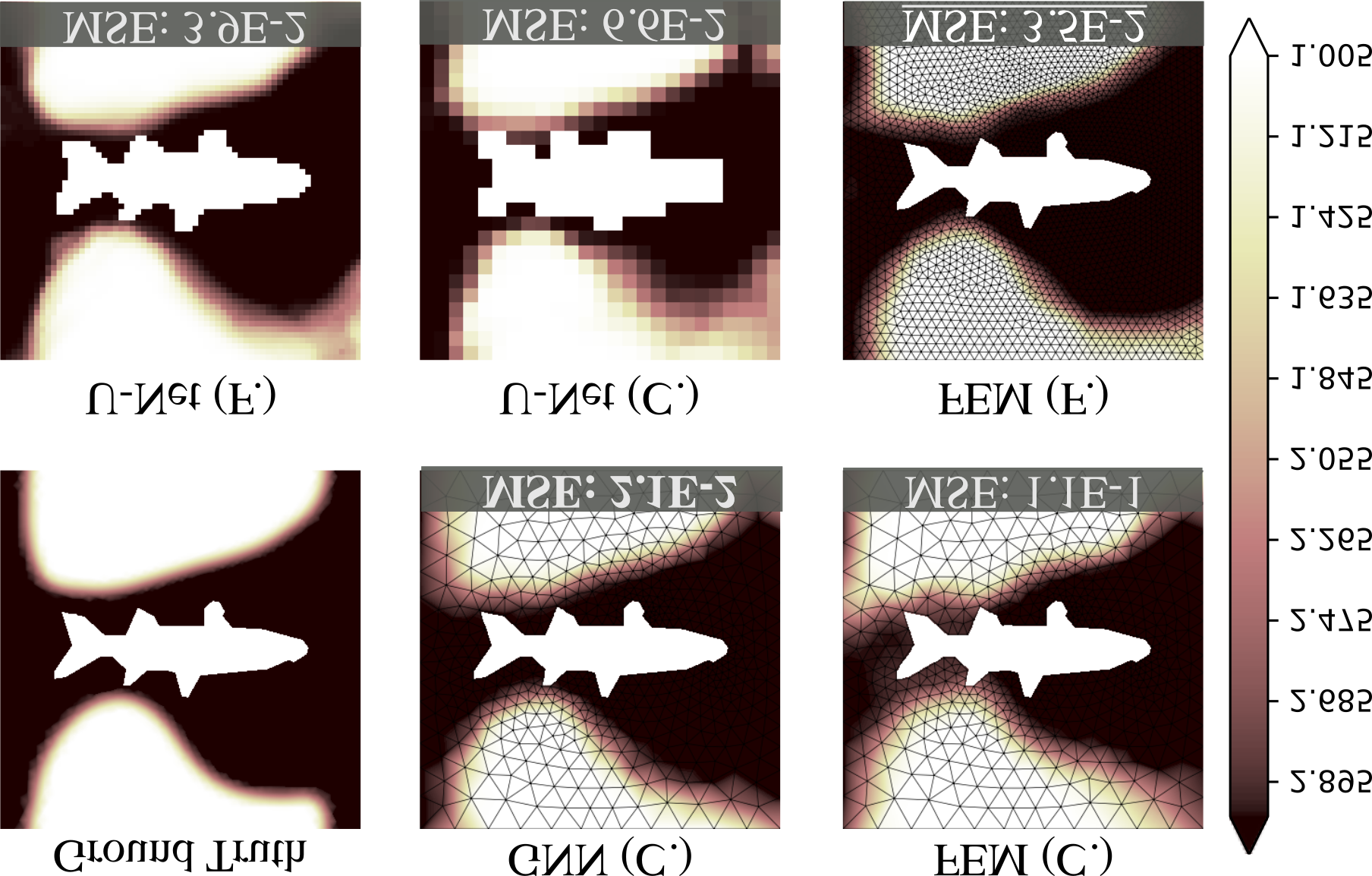}}
  \end{center}
  \caption{Qualitative results for full-waveform inversion experiments. Here ``C." refers to the coarse meshes and ``F." refers to the fine meshes, consistent with the notation used in \figref{wave_init_fig}. We observe that the GNN outperforms the U-Net (CNN) both quantitatively and qualitatively.}
  \label{fig:density_comparison}
\end{figure}

\Tableref{init_state_table} reports a quantitative comparison of the performance of different solvers used with or without the prior. We observe similar trends as in the initial state recovery task, where using a learned prior significantly improves the final accuracy. For this task, the proposed GNN solver gives the lowest MSE compared to the baselines. \Figref{density_comparison} shows qualitative comparisons between the different learned simulators and demonstrates that the proposed approach with the GNN forward model leads to a better recovery of the density distribution.\\

\begin{figure*}[ht!]
	\vspace{-0.2cm}
	\begin{center}
		\includegraphics[width=\textwidth]{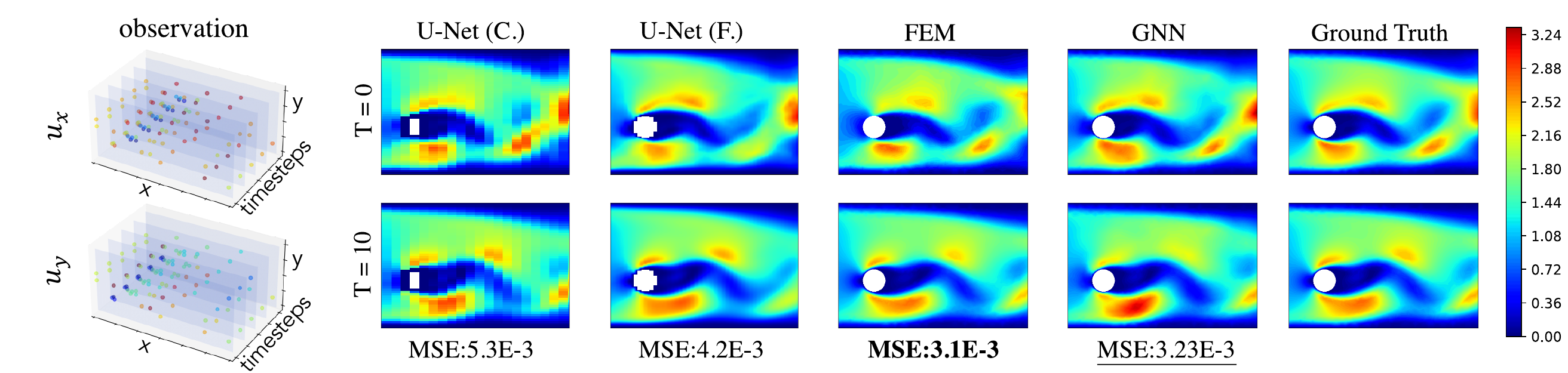}
	\end{center}
    \vspace{-0.3cm}
	\caption{Fluid data assimilation results using different forward models with the prior. Left: we have 50 sensors measuring the velocity field and we take measurements every 2 time steps over a flow clip of total 10 time steps. Right: qualitative comparison of recovered flows using different methods at $t=0$ (beginning of assimilation window) and $t=10$ (end of the assimilation window). We visualize the magnitude of the velocity ($\lVert \mathbf{u}\rVert^2_2$) for comparison (see supplementary for additional comparisons and results using different forward without the learned prior). The reported MSE are the average mean square error over the entire flow assimilation window. While the FEM solver yields the best results, it is roughly 90$\times$ slower than the GNN.} 
	\label{fig:fluid_demo_v}
\end{figure*}

\subsection{Two-Dimensional Incompressible Navier--Stokes Equation}
The two-dimensional incompressible Navier--Stokes Equations are non-linear partial differential equations modeling the dynamics of fluids. Taking a fluid density of $\rho=1.0$, the fluid is characterized by 
\begin{align}
	\frac{\partial \uxy}{\partial t} + \uxy\nabla\uxy - \nu \nabla^2 \uxy + \nabla p = 0\\
	\nabla \cdot \uxy = 0
\label{eq:ns}
\end{align}
Here, $\nu = 0.001$ is the kinematic viscosity, $\uxy$ represents the velocity of the fluid in the $x$ and $y$ directions, and $p$ represents the pressure. These incompressible Navier--Stokes equations can be solved given the initial condition $\uxy$ and boundary conditions for the given domain. For our experiments, we use an irregular domain $\Omega\in[0,1.6]\times [0,0.4]$ with a cylinder of random radius at a random position near the flow inlet. No-slip boundary conditions are defined at the lower and upper edge, as well as at the boundary of the cylinder as shown in \eqref{b1}. On the left edge, a constant inflow profile is prescribed, \eqref{inflow}; on the right edge, do-nothing boundary condition is used, \eqref{outflow}. 
\begin{align}
    \uxy(\x) = [0,0] \quad &\forall \x \in \partial \Omega_{\text{up, bottom, cylinder}}
    \label{eq:b1}\\ 
    \uxy(\x) = [1,0] \quad &\forall \x \in \partial \Omega_{\text{in}}
    \label{eq:inflow}\\
    \nu\partial_\eta \uxy - p\eta = 0 \quad &\forall \x \in \partial \Omega_{\text{out}}
    \label{eq:outflow}
\end{align}\\
We carry out flow assimilation on the velocity field $\uxy$ given sparse velocity measurements in space and time. Formally, we have
\begin{equation}
\begin{split}
    \minimize_{ \left\{ \latent \right\} }& \,\,\sum_{t\in T_{\text{iter}}}\|\measurementop(\uxy_t^{\text{gt}}(\x)) - \measurementop(\uxy_t^{\text{pred}}(\x)) \|_1 \quad \\
    s.t.\,\,&\uxy_{t+1} = \uxy_{t}+\evolutionop \left( \uxy_{t} \right) \\
            & \uxy_0(\x)=\prior_{\uxyinit}(z,\x).
\label{eq:ns_iv}
\end{split}
\end{equation}
where $\measurementop$ is the measurement sampling operator, and $\evolutionop$ is the forward model that solves the incompressible Navier--Stokes Equations.

Our assimilation trajectories are random flow clips of length T = 10 time intervals from the test dataset. We take measurements every 2 time intervals for a total of 5 measurement snapshots where each time intervals are equivalent to 1 learned solver timesteps, or 12.5 FEM solver timesteps. At every measurements, we have 50 sensors measuring the velocity field, $[u_x,u_y]$, as shown in \figref{fluid_demo_v}. \\
\begin{table}[h!]
\centering
\begin{tabular}{c|c|c|c}\toprule
\textbf{Forward Model} &\textbf{\# nodes} & \textbf{MSE} &\textbf{Runtime (s)}\\ \toprule
FEM  (Irr. C.)	   	& 2732  & \textbf{3.92e-4}& 3.87e1  \\
U-Net  (Reg. C.) 	& 2916  & 5.78e-3& \textbf{1.11e-1} \\
U-Net  (Reg. F.)	& 16384 & 2.55e-3& \underline{1.65e-1} \\
GNN  (Irr. C.)      & 2732  & \underline{9.73e-4}& 1.10e0  \\
\bottomrule
\end{tabular} 
\caption{Errors of different forward models for 50-steps rollout averaged over 50 test trajectories. The proposed method with the GNN provides better accuracy than the U-Net (CNN) at both coarse and fine resolution and is $35\times$ faster than the classical FEM solver.}
\label{table:solver_fluid}
\end{table}
\vspace{-0.4cm}
\paragraph{Learned Forward Simulation}
As described in Sec. \ref{sec:method}, our method uses a GNN as a fast and accurate learned simulator operating on irregular meshes. Similar to the 2D scalar wave equation experiment, we train as baselines a U-Net (CNN) for an input resolution of $54\times 54$, which matches the average number of nodes of the irregular grid used by the GNN ($\approx$2732 nodes).
We also train a U-Net (CNN) at $128\times 128$ resolution, which contains roughly $6\times$ more nodes compared to the coarse irregular grid (Irr. C.) used by the GNN. All learned simulators are trained in a supervised manner on a dataset obtained using an open source FEM solver ~\citep{LoggMardalEtAl2012}. Our dataset consists of 850 training trajectories on 55 meshes, and 50 test trajectories on 5 meshes. The unrolled MSE is shown in \Tableref{solver_fluid}. We observe a similar trend as in the wave equation example: the GNN gives the best accuracy among learned solvers and is approximately $35 \times$ faster than the FEM solver. 

\begin{figure}[t!]
	\begin{center}
		\includegraphics[width=0.38\textwidth]{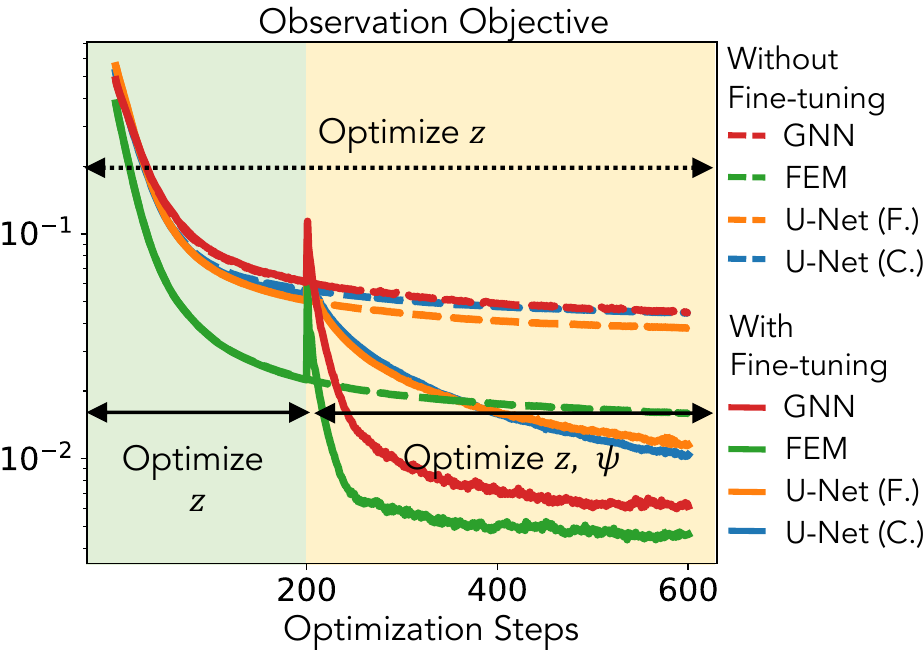}
	\end{center}
	\vspace{-0.2cm}
	\caption{Observation objective during the optimization. Solid lines: during the first 200 iterations, we only optimize the latent code $\latent$ and the prior, $\prior$,  parameterized by $\psi$ is fixed. From 200 - 600 iterations, we start to fine tune the prior, i.e optimizing $\psi$, to increase the expressiveness of the prior to better match the observations. After fine-tuning, we observe a significant decrease in the objective function. Dashed lines: we only optimize the latent code $\latent$ throughout the optimization. We observe that the final observation objective without fine-tuning is much higher than the cases with fine-tuning.} 
	\label{fig:fluid_demo_obj}
	\vspace{-0.2cm}
\end{figure}

\paragraph{Fluid Data Assimilation}
We train our prior $\prior_{\uxy}$ on 34000 fluid snapshots from the training trajectories, where $\prior_{\uxy}(\latent,\x) = \uxy (\x)= [u_x(\x), u_y(\x)]$. Similarly to the wave equation experiment, we solve the optimization problem  given in \eqref{ns_iv} using the ADAM optimizer \cite{kingma2014adam}. Different from the wave equation experiments we add a fine-tuning stage to the optimization where we update the parameters of the generative prior $\prior_{\uxy}$, which helps to improve the generalization performance in this problem. Similar techniques have been widely used in GAN inversion problems \cite{layer_opt,Pivotal}. \Figref{fluid_demo_obj} shows the averaged observation objective with optimization steps for different forward model and with or without fine-tuning procedure. After beginning the fine-tuning procedure we notice a spike in the objective function, but the optimization ultimately converges to much lower residuals than without the fine-tuning procedure.
\begin{table}[t!]
\centering
\begin{tabular}{l|ccc}\toprule
\textbf{Forward Model}	  & \multicolumn{3}{c}{\textbf{Fluid Data Assimilation}} \\
				& MSE ($u_x$) & MSE ($u_y$)  	  &  Runtime (s)			\\\midrule
FEM  (Irr. C.)  & {\textbf{4.46e-3}} & {\underline{1.94e-3}}		  & 6.39e1 	     \\
U-Net (Reg. C.)     & {1.02e-2}	    & {5.40e-3}		  & \textbf{1.27e-1}	 	      \\
U-Net (Reg. F.)     & {7.97e-3} 	& {3.31e-3}		  &	\underline{3.25e-1}			  \\
GNN  (Irr. C.)  & {\underline{6.55e-3}}& {\textbf{1.86e-3}}		  & 6.94e-1	  		  \\\bottomrule
\end{tabular}
\caption{Fluid assimilation results averaged over 50 test samples on 5 unseen meshes. Here, we simulate for 10 time steps taking measurement of 50 sampled sensors every 2 time steps. Runtime is the per-iteration optimization time. We observe that the GNN (Irr. C.) performs much better than the U-Net (Reg. C.) using a similar number of nodes (listed in \Tableref{solver_fluid}). U-Net (Irr. F.) uses $5\times$ more nodes than the GNN but has slightly worse performance. The FEM solver gives the best performance but is $90\times$ slower than the GNN.} 
\label{table:fluid_table}
\vspace{-0.5cm}
\end{table}
In \Tableref{fluid_table}, we qualitatively compare the data assimilation performance for different forward models with the learned prior, and qualitative comparisons can be found in \figref{fluid_demo_v}. We report the average MSE over all time steps of $u_x$ and $u_y$ as our evaluation metric. We observe similar trends as in the wave equation experiment: the learned GNN forward model with the learned prior achieves better accuracy compared to the coarse U-Net (CNN) model, which has a similar number of nodes. While the FEM solver gives the best fluid reconstruction, it is $50 \times$ slower than the GNN. U-Net (Reg. F.) uses $5\times$ more nodes than the GNN, but achieves similar performance.


\section{Discussions}
\label{sec:discussions}
In this paper we present a general framework for solving PDE-constrained inverse problems with a GNN-based forward solver and an autodecoder-style learned prior. Both the GNN-based learned simulator and the learned prior operate on irregular meshes with adaptive resolution, enabling representing and processing signals with fewer nodes compared to the regular grids used by convolutional networks. Our experiments on the wave equation and Navier--Stokes equations demonstrate that the proposed framework achieves improved performance compared to a conventional U-Net (CNN) operating on grids with the same number of nodes as the GNN. Moreover, our approach is up to $90\times$ faster compared to conventional FEM solvers.

We also mention a few limitations of the proposed approach. The current implementation is limited in the number of time-steps that can be modeled in the inverse problem due to memory required to unroll the solvers.
Also, the method requires the forward model to be trained from scratch for changes to the underlying equations (e.g., significant changes to the boundary conditions or dynamics).
Still, techniques like gradient checkpointing~\cite{Checkpointing} may be helpful to mitigate the memory constraints of the unrolled optimization, and generalizing learned physics solvers across problem settings is an promising direction for future work.

Overall, our approach works to integrate GNNs and learned priors for solving physics-constrained inverse problems.
Our approach may be useful for solving ill-posed problems across a range of tasks related to physics-based simulation and modeling. 
GNNs are an attractive architecture for efficient modeling across multi-resolution domains and may yield even greater improvements compared to conventional CNNs, especially for larger domains or in 3D problem settings.

\section{Acknowledgements}
\label{sec:acknowledgements}
This project was supported in part by a PECASE by the ARO and Stanford HAI.

\bibliography{references}
\bibliographystyle{icml2022}

\pagebreak
\onecolumn
\section*{Supplementray}
\beginsupplement
\label{sec:Supplementray}

\section{Two-Dimensional Scalar Wave Equation}
\subsection{Dataset}
The dataset for the wave equation consists of 1100 training trajectories over 37 meshes, and 40 test trajectories over 3 unseen meshes. For each mesh, we have a fish shape obstacle at the center obtained from the shape dataset \citesupp{shape_dataset}. The Dirichlet boundary condition is applied to all boundaries.  The ground truth trajectories are obtained using an open source FEM solver, FEniCs \citesupp{LoggMardalEtAl2012}, simulated on fine meshes with Euler method and first order elements. We use generalized minimal residual method (GMRES) as our linear solver and incomplete LU factorization as our preconditioner. The supervised fields on coarse meshes and regular meshes are all interpolated from the ground truth trajectories. For every trajectory, we randomly sample an initial wavefield, and a velocity distribution from the gaussian random field. The velocity distribution is threshold to binary. To satisfy the boundary conditions, the sampled initial wavefield is decayed to 0 near the boundaries using the solution of the Eikonal equation.
\subsection{Prior Network}
For the prior networks, $\prior_c$ or $\prior_u$, we use a multilayer perceptron (MLP) with 6 hidden layers, hidden features of size $256$, and the positional encoding \citesupp{tancik2020fourier}. ReLU nonlinearity is used at all intermediate layers and sigmoid is used at the final layer. Input latent code is of size $64$. During the training, the dataset is normalized to be between $[0,1]$. We train the network with ADAM optimizer \citesupp{kingma2014adam} for 1500 epochs using a batch size of 32, and a learning rate of 5e-4. For each training sample, we randomly sample 900 points and add random noise to the coordinates for better generalization. The loss function is \eqref{loss_prior}, where $\sigma=0.01$. 

\begin{align}
    \argmin_{\psi_\prior,z}\sum_{i=1}^N \left(\sum_{j=1}^{K}\loss \left( \prior \left(z^i,\x_j\right),u^i_j \right) +\frac{1}{\sigma^2} \|z^i\|_2 ^2\right)
\label{eq:loss_prior}
\vspace{-0.1cm}
\end{align}

\subsection{Learned Simulators}
\paragraph{U-Net}
For the U-Net learned forward model, we adopt the architecture in \citesupp{thuerey:2020}. Input of the network consists of 3 channels, $[u,u', \text{mask}]$, where the "mask" indicates the simulation domain. We adjust the network size so that the total number of parameters of the U-Net matches the total number of parameters used in the GNN. The input field are all normalized to zero mean and unit variance using the dataset statistics. We train the network for 500 epochs using the ADAM optimizer \citesupp{kingma2014adam}, learning rate 0.0004 and batch size 10.
\paragraph{GNN}
For GNN, we adopt the architecture in \citesupp{pfaff2020learning}. The input edge features consist of the relative displacement vector between nodes and its norm. The input node features consist of $[u, u',c, \text{node type}]$, where node type is a one-hot vector that is $[0,1]$ on the boundaries, or $[1,0]$ off boundaries. We use two-hidden-layer MLPs with hidden features size $256$, and the total number of message passing steps is 10. The input and the output features are all normalized to zero mean and unit variance using the dataset statistics. We train the network using the ADAM optimizer \citesupp{kingma2014adam} with a learning rate decay exponentially from 1-4 to 1e-8 over 500 epochs. 
\subsection{Qualitative Comparisons}
\begin{figure}[h]
    \centering
    \includegraphics[width=\textwidth]{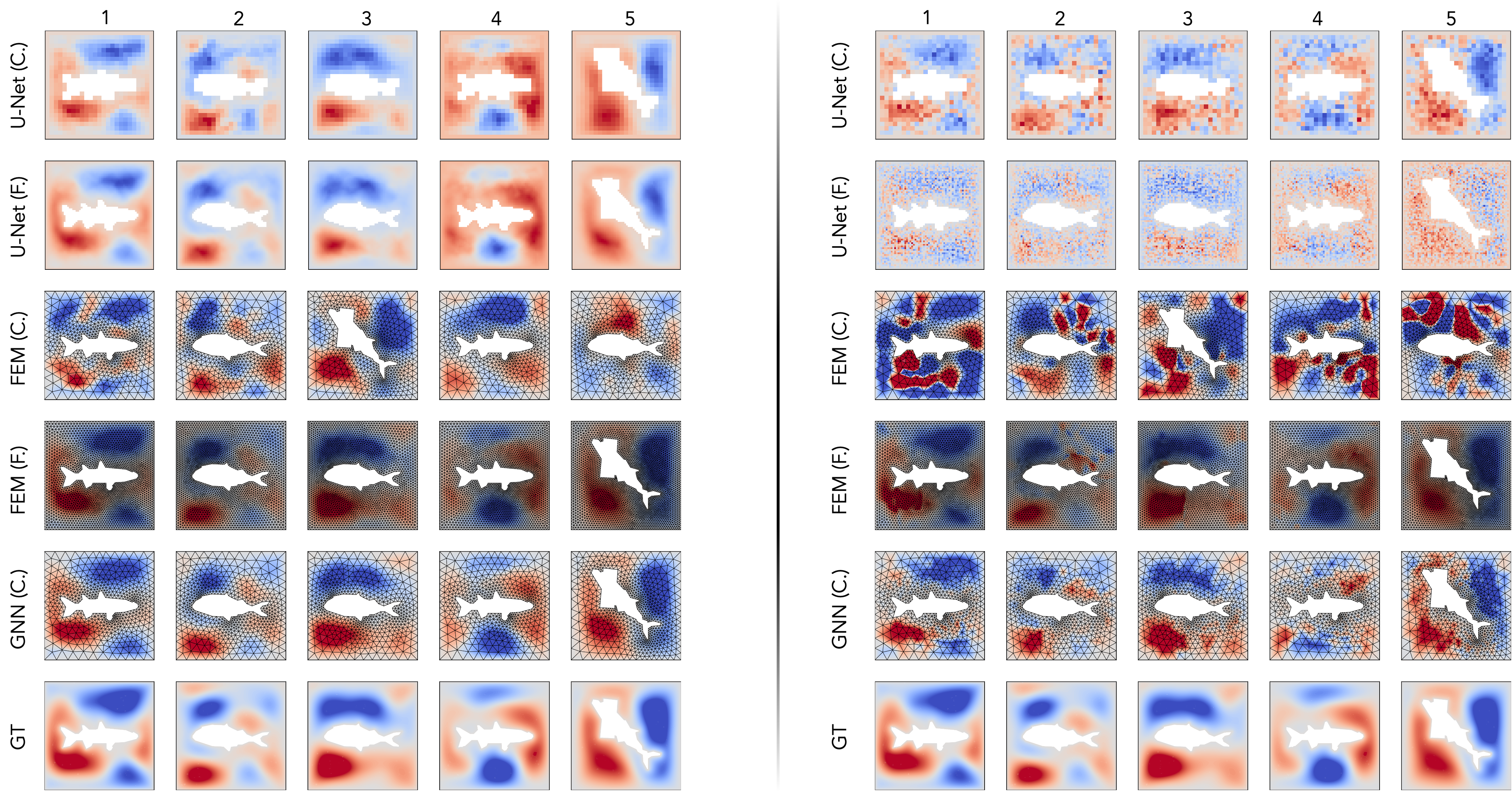}
    \caption{Sampled qualitative comparisons for initial state recovery results with the wave equation. Left: with the learned prior; Right: without the learned prior. We observe that with prior gives much better results than without prior in all cases, indicating the effectiveness of our prior.}
    \label{fig:wave_compare_init}
\end{figure}

\begin{figure}[h]
    \centering
    \includegraphics[width=\textwidth]{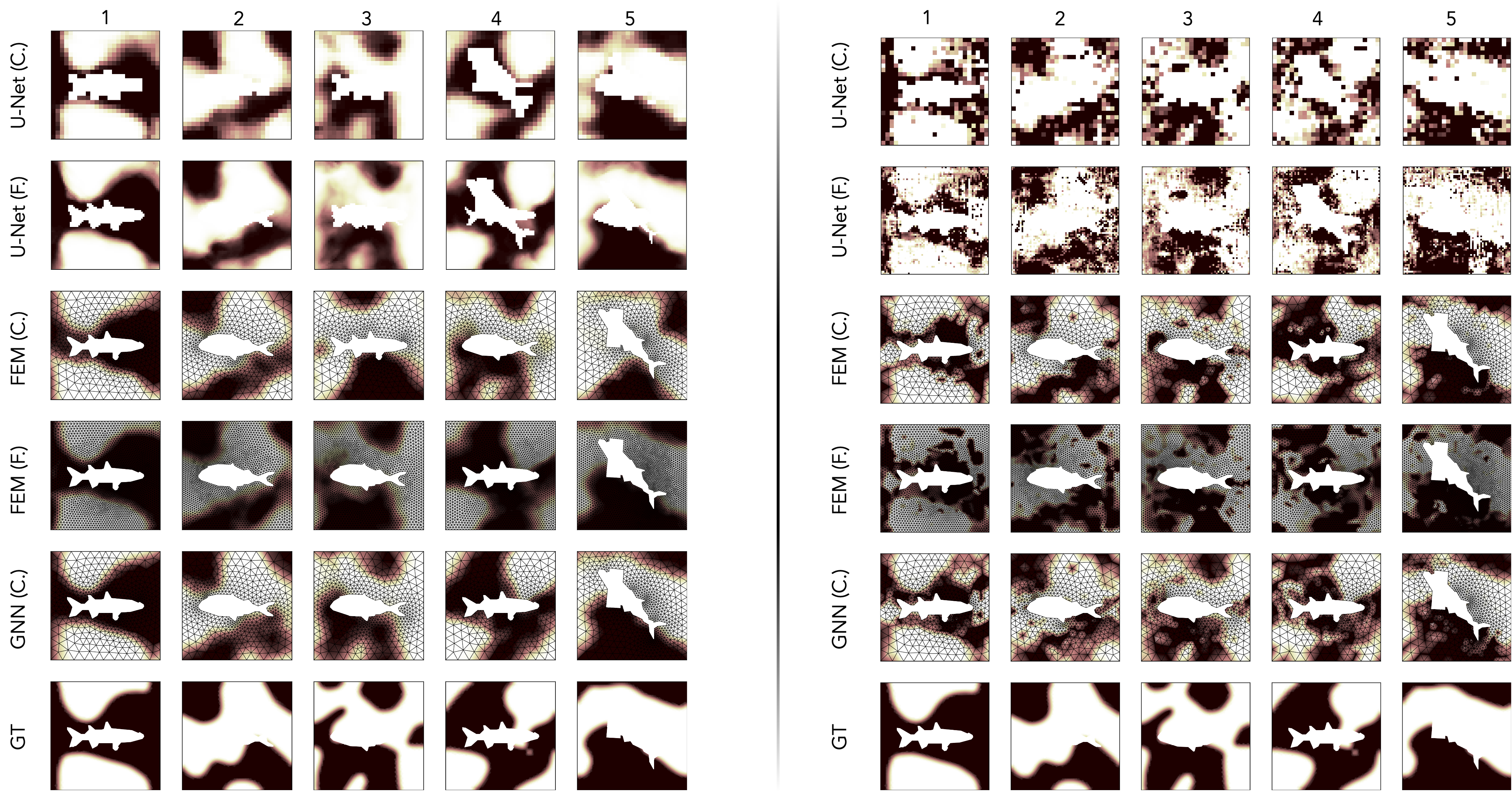}
    \caption{Sampled qualitative comparisons for full-waveform inversion results with the wave equation. Left: with the learned prior; Right: without the learned prior. We observe that with prior gives much better results than without prior in all cases, indicating the effectiveness of our prior.}
    \label{fig:wave_compare_density}
\end{figure}

Here, we show some qualitative comparisons for the initial state recovery, \figref{wave_compare_init}, and the full-waveform inversion task, \figref{wave_compare_density}, for the 2D wave equation experiments discussed in Sec. 4.1. In these experiments, the sensor locations are off boundary nodes randomly sampled from the coarse irregular grid that the GNN operates on. 
\clearpage
\newpage
\section{Two-Dimensional Incompressible Navier Stokes
Equation}
\subsection{Dataset}
The fluid dataset consists of 850 training trajectories on 55 meshes, and 50 test trajectories on 5 meshes. The ground truth trajectories are obtained through simulation using an open source FEM solver, FEniCs \citesupp{LoggMardalEtAl2012} on the fine meshes and Chorin’s method \citesupp{Chorin}. For every trajectory, we randomly sample an initial velocity, $\uxy$, from a gaussian random field. The first 1000 steps of the simulation are discarded. 
\subsection{Prior Network}
For the prior networks, $\prior_\uxy$, we use a multilayer perceptron (MLP) with 6 hidden layers, hidden features of size $128$, and the positional encoding \citesupp{tancik2020fourier}. ReLU nonlinearity is used at all intermediate layers. Input latent code is of size $128$. The skip connection is adopted here, i.e. the input latent code is concatenated to the 3rd and the 5th layer of the MLP \citesupp{park2019deepsdf}. During the training, the dataset is normalized to be zero mean and unit variance. We train the network with the ADAM optimizer \citesupp{kingma2014adam} for 1500 epochs using a batch size of 32 and a learning rate of 5e-4. For each training sample, we randomly sample 4000 points and add random noise to the coordinates for better generalization. The loss function is \eqref{loss_prior}, where $\sigma=0.01$. 
\subsection{Learned Simulators}
\paragraph{U-Net}
For the U-Net learned forward model, we adopt the architecture in \citesupp{thuerey:2020}. The input of the network consists of 3 channels, $[u_x,u_y,\text{mask}]$, where the "mask" indicates the simulation domain. We adjust the network size so that the total number of parameters of the U-Net matches the total number of parameters used in the GNN. We train the model for 500 epochs with the ADAM
optimizer \citesupp{kingma2014adam}, a learning rate of 0.0004 and a batch size of 128. 
\paragraph{GNN}
For the GNN learned forward model, we adopt the architecture in \citesupp{pfaff2020learning}. The input edge features consist of the relative displacement vector between nodes, and its norm. The input node features consist of $[u_x,u_y,\text{node type}]$, where the node type is a one-hot vector indicates if the node is at the inlet, outlet, cylinder boundary, wall boundaries, or otherwise. We use two-hidden-layers MLPs with hidden features size $128$, and the total number of the message passing steps is 15. We train the network with the ADAM optimizer \citesupp{kingma2014adam} with a learning rate decay exponentially from 1-4 to 1e-8 over 500 epochs. 
\subsection{Fine tuning}
\begin{table}[h]
\centering
\begin{tabular}{l|cccc}\toprule
\textbf{Forward Model}	  & \multicolumn{4}{c}{\textbf{Fluid Data Assimilation}} \\
                          & \multicolumn{2}{c}{\textbf{without Fine-tuning}} & \multicolumn{2}{c}{\textbf{with Fine-tuning}} \\
                          & MSE ($V_x$)      & MSE ($V_y$)           & MSE ($V_x$) & MSE ($V_y$) 		\\\midrule
FEM  (Irr. C.)            & \textbf{5.58e-3} & \textbf{2.83e-3}    & \textbf{4.46e-3} & \underline{1.94e-3} \\
U-Net (Reg. C.)           & 1.21e-2   & 6.17e-3                      & {1.02e-3}	    & {5.40e-3}	 \\
U-Net (Reg. F.)           & \underline{8.64e-3}   & \underline{4.04e-3}                      & {7.974e-3} 	& {3.31e-3} \\
GNN  (Irr. C.)            & 9.87e-3   & 4.19e-3   & {\underline{6.55e-3}}& {\textbf{1.86e-3}}\\\bottomrule
\end{tabular}
\label{table:fluid}
\caption{Fluid assimilation results averaged over 40 test samples on 5 unseen meshes. Here, each flow clip consists of 10 time steps, where each time step is equivalent to 2 learned solver steps (GNN or U-Net) and 12.5 FEM solver step. We take measurement of the velocity, $[u_x,u_y]$, every 2 time steps using 50 sampled sensors. Here we compare the with-fine-tuning and the without-fine-tuning approach. We observe that with-fine-tuning approach achieves lower MSE for all forward models.} 
\end{table}
\subsection{Qualitative Comparisons}
\begin{figure}[h!]
    \centering
    \includegraphics[width=0.9\textwidth]{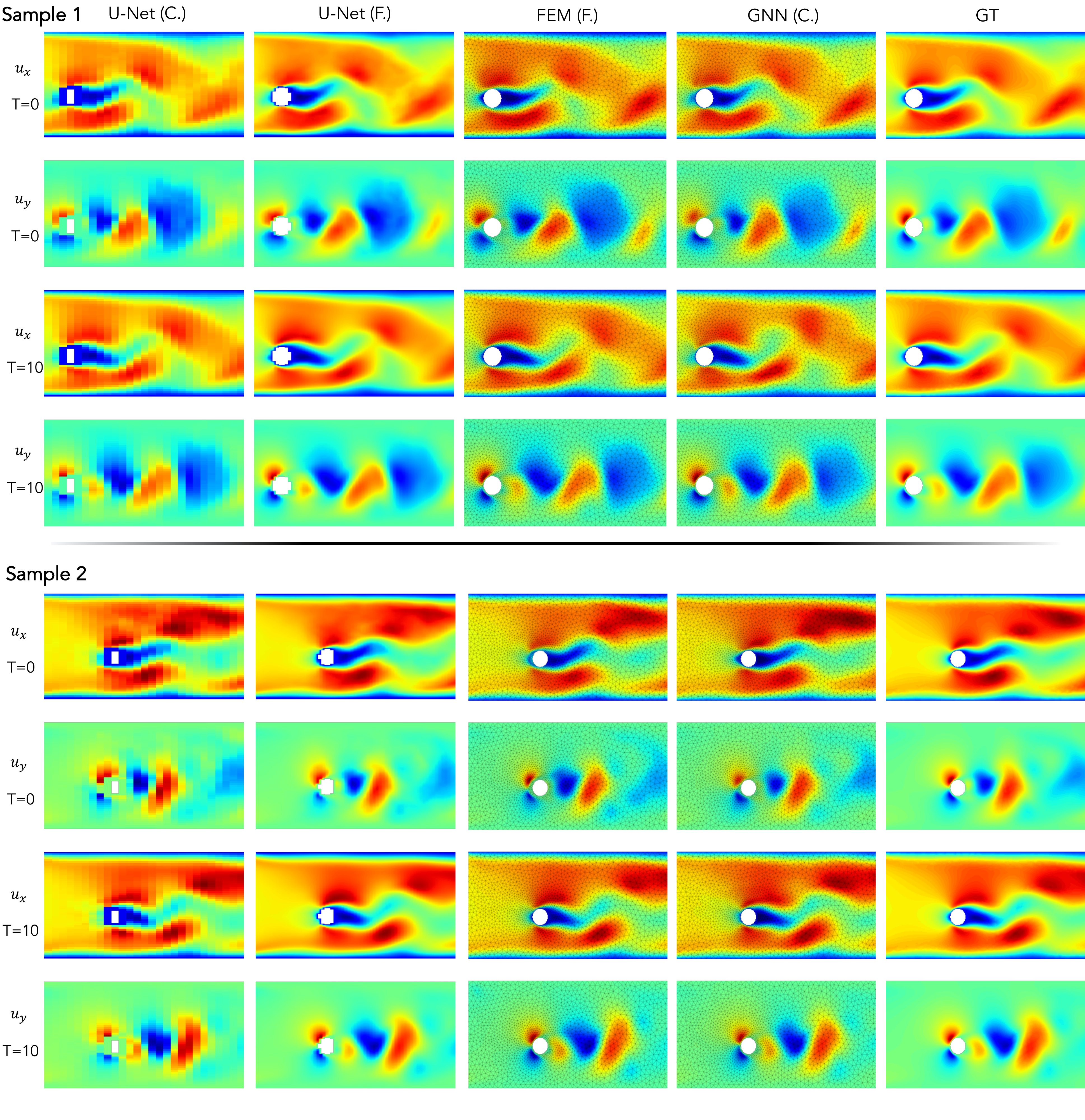}
    \caption{Sample comparisons for flow assimilation task.}
    \label{fig:fluid_compare}
\end{figure}

In \figref{fluid_compare}, we show some qualitative comparisons for the flow assimilation task using different solvers, and learned prior. In this experiment, 10 sensors are sampled at a bounding box near the cylinder, and 40 sensors are sampled randomly from the whole domain. 

\subsection{Without Prior}
\begin{figure}[h]
  \centering
  \includegraphics[width=0.9\textwidth]{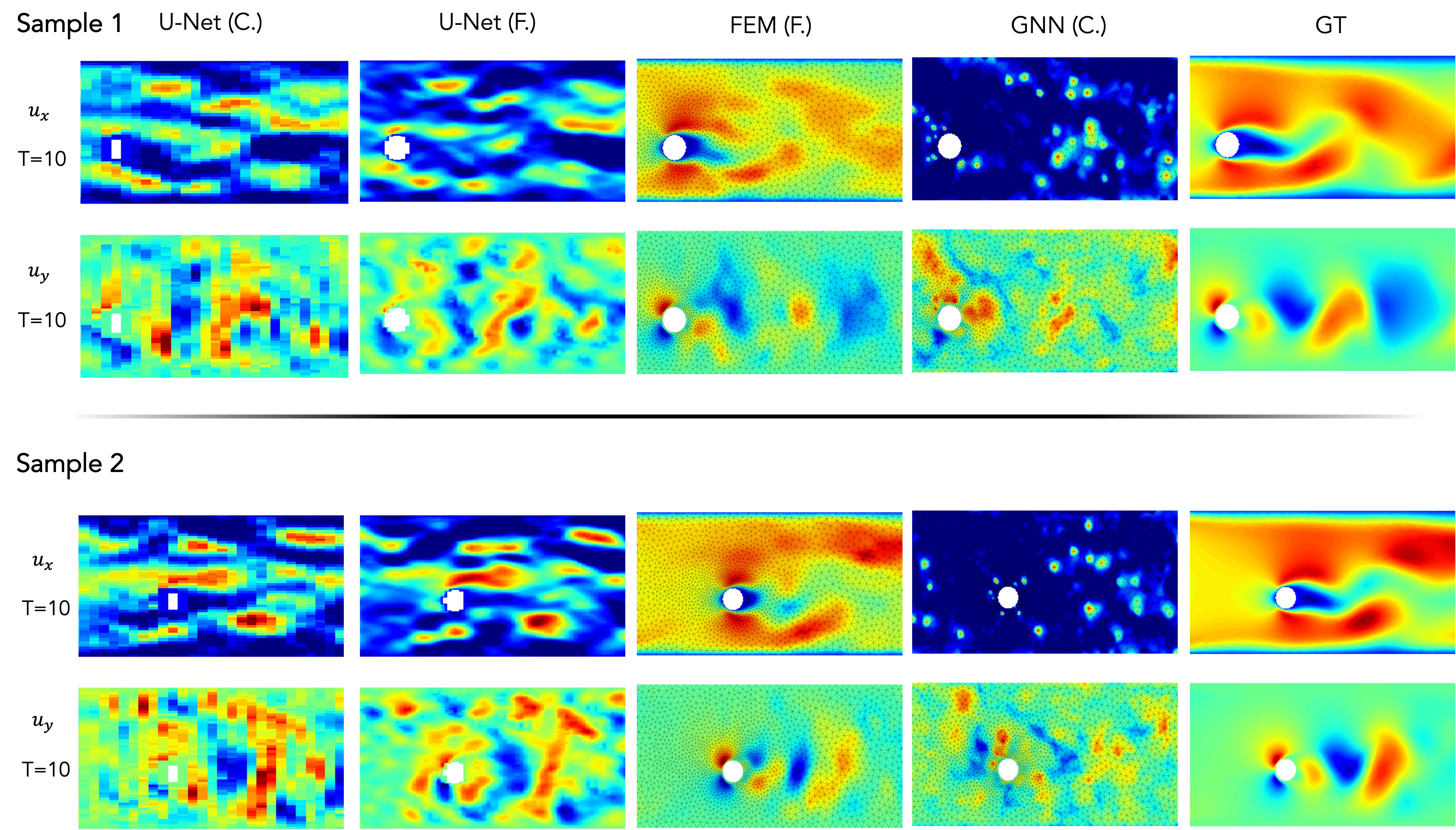}
  \caption{Sample comparisons for flow assimilation task without using learned prior network.}
  \label{fig:fluid_compare_np}
\end{figure}

In \figref{fluid_compare_np} and \tableref{fluid_np}, we show some qualitative and quantitative comparisons for the flow assimilation task without using the learned prior network. Due to the highly ill-posedness of the problem, all solvers yield results deviates from the ground truth. All learned solvers yield results with strong artifacts as the input flow velocity is now far outside the training dataset distribution that the learn solvers trained on. Note that for the learned simulators, the learned prior network also ensures that the input states/physics parameters are always within the training set distribution.
\begin{table}[h!]
  \centering
  \begin{tabular}{l|cc}\toprule
  \textbf{Forward Model}	  &MSE ($V_x$)      & MSE ($V_y$)     	\\\midrule
  FEM  (Irr. C.)          & 2.45e-1 & 3.13e-2 \\
  U-Net (Reg. C.)         & 8.50e-1 &  5.44e-2             \\
  U-Net (Reg. F.)           & 9.76e-1    & 5.55e-2     \\
  GNN  (Irr. C.)            & 1.97e0 & 4.85e-2 \\\bottomrule
  \end{tabular}
  \label{table:fluid_np}
  \caption{Fluid assimilation results without using the learned prior network.} 
\end{table}

\section{Runtime Analysis}
To obtained the average runtime per optimization iteration, we run our approaches on 8 CPU cores (FEM solver) and a single Quadro RTX 6000 GPU (U-Net and GNN).

\bibliographysupp{references}
\bibliographystylesupp{icml2022}
\end{document}